\documentclass[conference]{IEEEtran}
\usepackage{graphicx, epsfig,amsmath,amssymb,latexsym,graphics,float}
\usepackage{cite}
\usepackage{setspace,subfig}
\usepackage{lipsum}
\usepackage{multicol}
\usepackage{bbding}
\begin{document}
\begin{onecolumn}
\thispagestyle{empty}
\vspace*{0.6cm}
\begin{center}

\vspace{-.2in}

{\LARGE Vision-based Traffic Flow Prediction using Dynamic Texture Model and Gaussian Process
}\\


{\large \vspace{1cm}
\vspace{0.3cm}

Bin Liu\footnote{Corresponding author.
E-mail: {\sf bins@ieee.org}.\\}, Hao Ji and Yi Dai \vspace{0.5cm}

{$~$School of Computer Science and Technology, \\Nanjing University of Posts and Telecommunications}\\
{$~$Nanjing, Jiangsu, 210023, China\\}

\vspace{1cm}

Manuscript Submitted --- July 14th, 2016\\
\vspace{1cm}
This Manuscript has been accepted by \\2017 the Second International Conference on Multimedia and Image Processing (ICMIP 2017)\\
} 

\end{center}

\end{onecolumn}
\newpage
\begin{twocolumn}
\title{Vision-based Traffic Flow Prediction using Dynamic Texture Model and Gaussian Process}
\author{\IEEEauthorblockN{Bin Liu$^($\Envelope $^)$}
\IEEEauthorblockA{School of Computer Science and Technology\\
Nanjing University of Posts and Telecommunications\\
Nanjing, Jiangsu 210023, China\\
Email: bins@ieee.org}
\and
\IEEEauthorblockN{Hao Ji and Yi Dai}
\IEEEauthorblockA{School of Computer Science and Technology\\
Nanjing University of Posts and Telecommunications\\
Nanjing, Jiangsu 210023, China}}
\maketitle
\begin{abstract}
In this paper, we describe work in progress towards a real-time vision-based traffic flow prediction (TFP) system.
The proposed method consists of three elemental operators, that are dynamic texture model based motion segmentation, feature extraction and Gaussian process (GP) regression. The objective of motion segmentation is to recognize the target regions covering the moving vehicles in the sequence of visual processes. The feature extraction operator aims to extract useful features from the target regions. The extracted features are then mapped to the number of vehicles through the operator of GP regression. A training stage using historical visual data is required for determining the parameter values of the GP.
Using a low-resolution visual data set, we performed preliminary evaluations on the performance of the proposed method. The results show that our method beats a benchmark solution based on Gaussian mixture model, and has the potential to be developed into qualified and practical solutions to real-time TFP.
\end{abstract}
\IEEEpeerreviewmaketitle
\section{Introduction}
\label{sec:intro}
As one of the most critical issues to implement an intelligent transportation system (ITS), real-time traffic flow prediction (TFP) has gained more and more attentions in recent years. The objective of TFP is to provide traffic flow information that has the potential to help road users make better decisions on traveling, improve traffic operation efficiency, alleviate traffic congestion and reduce carbon emissions \cite{lv2015traffic}. With the rapid development and deployment of various sensor sources, such as inductive loops \cite{jeng2014high}, radars \cite{felguera2012vehicular}, visual sensors \cite{sun2004road,sun2006road}, mobile global positioning systems \cite{zhang2013aggregating}, crowd sourcing \cite{schnitzler2014heterogeneous} and social media \cite{pan2013crowd}, traffic data are exploding. Actually we have
entered the era of data driven traffic prediction.

Although there have been already many TFP systems and models, how to implement accurate, practical and economical TFP is still a challenging subject. For this reason, it is always useful to explore new principles and approaches that allow wide-spread real applications. In this spirit, we are dedicated to developing an easy to implement real-time vision-based TFP system.

Here we describe work in progress in using dynamic texture model \cite{chan2008modeling} and Gaussian process (GP) \cite{rasmussen2006gaussian} to do real-time vision-based TFP.
A conceptual scheme of our method is presented in Fig.\ref{fig:framework}. The proposition of this TFP scheme is inspired by the success of the application of the dynamic texture model and GP in counting pedestrians with low-level visual features \cite{chan2012counting}.
\begin{figure*}
\begin{tabular}{c}
\centerline{\includegraphics[width=5.5in,height=4.5in]{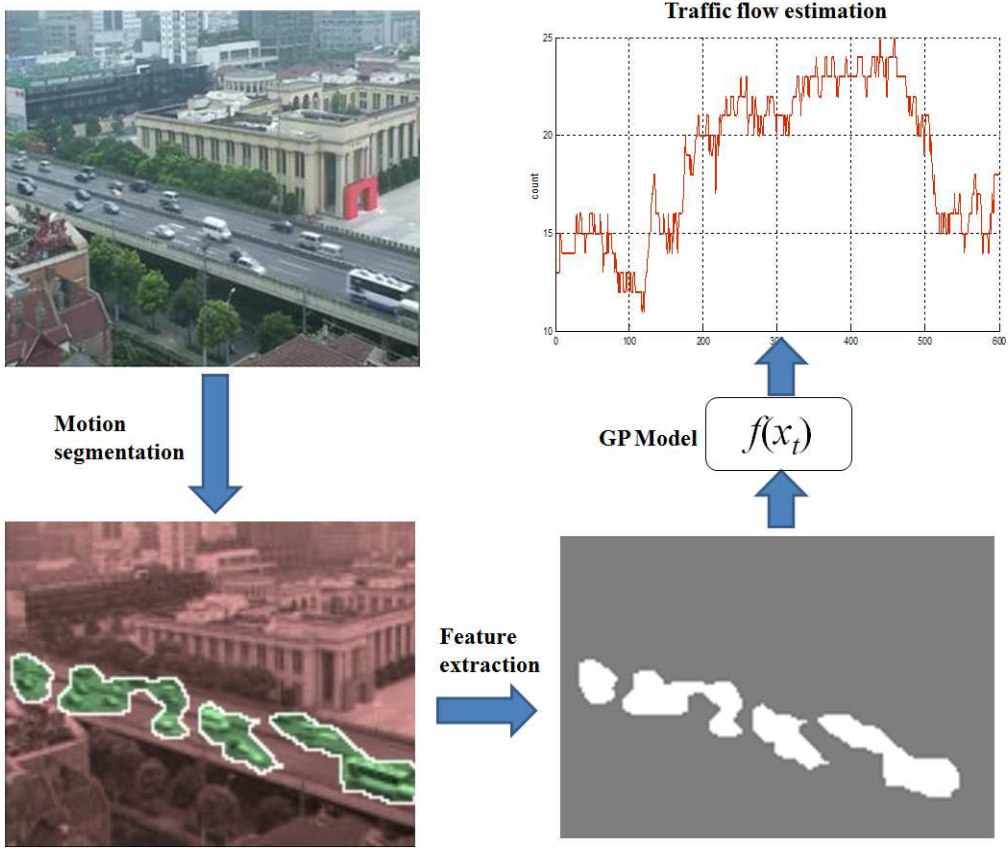}}
\end{tabular}
\caption{Traffic flow prediction system: the scene is segmented and the motion area is recognized.
Features that account for the motion area are extracted, and the traffic flow is estimated with a Gaussian process.}\label{fig:framework}
\end{figure*}
Preliminary evaluation results using low-resolution visual data demonstrate that this scheme has potentials to be developed into qualified solutions to the real-time TFP problem.
\section{The proposed TFP scheme}
A graphical illustration of the proposed TFP scheme is shown in Fig.\ref{fig:framework}. This scheme is mainly composed of three operators, that are the dynamic texture model based motion segmentation, feature extraction and GP regression. In what follows, we briefly introduce the necessary details of each operator.
\subsection{Dynamic texture model based motion segmentation}
The dynamic texture model is a suitable tool for representing dynamics in visual processes \cite{chan2008modeling}.
We adopt this model here for decomposing a visual sequence into a statistically spatio-temporal homogeneous target region and a static background region. The target region is hoped to be able to cover all the moving vehicles in the visual scene, and thus is the object to be processed by the follow-up feature extraction and GP regression operators, which will be described in Subsections \ref{sec:feature} and \ref{sec:GP}, respectively.
The task of motion segmentation then turns to be how to recognize the target region online. The success of motion segmentation depends on the ability of the dynamic texture model to capture the spatio-temporal law of the moving vehicles. This law is expected to be embedded in some statistically homogeneous features.

A dynamic texture model consists of a \emph{hidden state variable} $x_t$, and an \emph{observed variable} $y_t$, which are related through a linear dynamical system defined by
\begin{equation}
\left\{\begin{array}{ll}\label{eqn:lds}
x_{t+1}=Fx_t+w_t \\
y_t=Hx_t+v_t \end{array} \right.
\end{equation}
where $x_t\in\mathbb{R}^n$ and $y_t\in\mathbb{R}^m$. The parameter $F\in\mathbb{R}^{n\times n}$ denotes the \emph{state transition matrix} and $H\in\mathbb{R}^{m\times n}$ is the o\emph{bservation matrix}. The driving noise process $w_t$ is a zero mean Gaussian distribution with covariance $Q$, i.e., $w_t\sim\mathcal{N}(0,Q)$, where $Q$ is a positive-definite $n\times n$ matrix. The observation noise $v_t$ is also Gaussian distributed with zero mean and covariance $R$, i.e., $v_t\sim\mathcal{N}(0,R)$. The initial state $x_1$ is allowed to have arbitrary mean and covariance, i.e., $x_1\sim\mathcal{N}(\mu,P)$.

During the motion segmentation process, the dynamic texture model is first learned, and the video is then segmented by assigning locations of the moving vehicles to the dynamic texture model. The moving law of the vehicles is captured by the state transition function in Eqn.(\ref{eqn:lds}). In the model learning stage, the video is first represented as a bag of spatio-temporal patches, which is then clustered using the Expectation-Maximization algorithm \cite{dempster1977maximum}. For more details on dynamic texture modeling based motion segmentation, readers are referred to \cite{chan2008modeling}.
\subsection{Feature extraction}\label{sec:feature}
This operator aims to extract features to capture segment properties such as shape and size. The extracted features constitute the input of the GP, which maps them to the count of vehicles in the scene. The basic assumption adopted here is that there exist certain linear or nonlinear relationships between the visual features and the number of vehicles in the scene.
The above assumption is reasonable considering that low-level visual features have already been successfully used to predict the number of people \cite{chan2012counting}.

We select features used in \cite{chan2012counting} for capturing the properties of the target region, which is segmented out of the scene by the operator of feature extraction. The features we are concerned with include:
\begin{itemize}
\item \emph{Area}, namely the number of pixels in the target region.
\item \emph{Perimeter}, i.e., the number of pixels on the perimeter of the target region.
\item \emph{Perimeter-area ratio}, namely the ratio between the perimeter and the area of the target region.
\item \emph{Total edge pixels}, which represent the total number of edge pixels contained in the target region.
\item \emph{Homogeneity}, which describes the texture smoothness and is defined to be $g_{\theta}=\sum_{i,j}p(i,j|\theta)/(1+|i-j|)$.
\item \emph{Energy}, which is defined to be $e_{\theta}=\sum_{i,j}p(i,j|\theta)^2$.
\item \emph{Entropy}, which describes the randomness of the texture distribution and is defined to be $h_{\theta}=\sum_{i,j}p(i,j|\theta)\log p(i,j|\theta)$.
\end{itemize}
In the definitions of the last three features, the calculations are based on the images that have been quantized into eight
gray levels and masked by the motion segmentation process. Specifically, $p(i,j|\theta)$ denotes the joint probability of neighboring pixel values and is estimated for four orientations, namely $\theta\in\{0^{\circ}, 45^{\circ}, 90^{\circ}, 135^{\circ}\}$. A set of three features is extracted for each $\theta$ for a total of 12 texture features. See more details in \cite{chan2012counting}.
\subsection{GP regression}\label{sec:GP}
GPs are useful tools for Bayesian machine learning \cite{rasmussen2006gaussian}.
The GP regression is a procedure of Bayesian inference of continuous values, wherein a GP is used as a prior probability distribution over functions.
In GP regression, it is assumed that for a Gaussian process $f$ observed at coordinates $x$, the vector of values $f(x)$ is just one sample from a multivariate Gaussian distribution of dimension equal to the number of observed coordinates $|x|$. Therefore, under the assumption of a zero-mean distribution, $f(x)\sim \mathcal{N}(\textbf{0},K(\theta,x,x'))$, where $K(\theta,x,x')$ is the covariance matrix between all possible pairs $(x,x')$ for a given set of hyperparameters $\theta$ \cite{rasmussen2006gaussian}. The value of $\theta$ is determined through maximizing the log marginal likelihood:
\begin{eqnarray}
\log p(f(x)|\theta,x)&=&-\frac{1}{2}f(x)^TK(\theta,x,x')^{-1}f(x)\\
& &-\frac{1}{2}\log\det(K(\theta,x,x'))-\frac{|x|}{2}\log2\pi, \nonumber
\end{eqnarray}
where $A^T$ denotes the transposition of $A$. Given the specified value of $\theta$, we can make predictions about unobserved values $f(x^{\ast})$ at coordinates $x^{\ast}$ by drawing samples from the predictive distribution $p(y^{\ast}|x^{\ast},f(x),x)=\mathcal{N}(y^{\ast}|A,B)$ where the posterior mean estimate $A$ is defined as
\begin{equation}
A=K(\theta,x^{\ast},x)K(\theta,x,x')^{-1}f(x)
\end{equation}
and the posterior variance estimate $B$ is defined as:
\begin{equation}
B=K(\theta,x^{\ast},x^{\ast})-K(\theta,x^{\ast},x)K(\theta,x,x')^{-1}K(\theta,x^{\ast},x)^T
\end{equation}
where $K(\theta,x^{\ast},x)$ is the covariance between the new coordinate of estimation $x^{\ast}$ and all other observed coordinates $x$ for a given hyperparameter vector $\theta$. $K(\theta,x^{\ast},x^{\ast})$ is the variance at point $x^{\ast}$ as dictated by $\theta$.
Note that the classes of functions that the GP
can model is dependent on the kernel covariance function used. Here we adapt a kernel function that was previously used for the
task of pedestrian counting for the task of TFP. This function is defined to be \cite{chan2008privacy}:
\begin{equation}
k(x_r,x_s)=\beta_1(x_r^Tx_s+1)+\beta_2\exp(\frac{-\|x_r-x_s\|^2}{\beta_3})+\beta_4\delta(r-s)
\end{equation}
where $\|x-y\|$ denotes the Euclidean distance between $x$ and $y$, $\delta(\cdot)$ is the dirac delta function and $\beta=\{\beta_1,\beta_2,\beta_3,\beta_4\}$ denotes the hyperparameters.
\section{Experimental evaluation}
Here we report preliminary results in applying the proposed scheme to analyze real visual traffic data. The whole data set consist of 1000 frames of images and it assumes that the camera is stationary. We first convert the video into a sequence of images and then resize them so that they have a fixed 160 rows and 110 columns. In Fig.\ref{traffic}, we show two examples of the image frames after the above preprocessing.
\begin{figure*}
\begin{tabular}{c}
\centerline{\includegraphics[width=5in,height=2in]{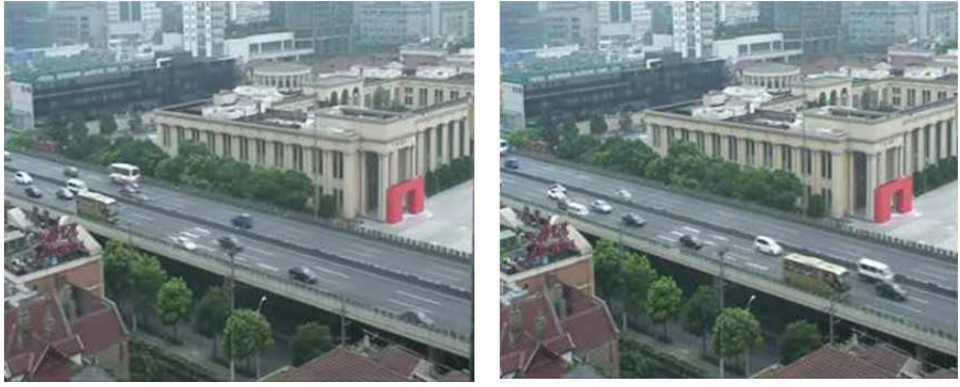}}
\end{tabular}
\caption{An example show of two frames of images in the experimental visual sequences.}\label{traffic}
\end{figure*}
%

We splitted the sequence of images with $60\%$ of them reserved for training and the remaining $40\%$ for testing.
See Fig.\ref{fig03} for the estimation result given by the GP regression. It is shown that, despite that the images under analysis
are very low-resolution, the GP still gives a rough estimate of the number of vehicles online.
\begin{figure*}
\begin{tabular}{c}
\centerline{\includegraphics[width=4in,height=2in]{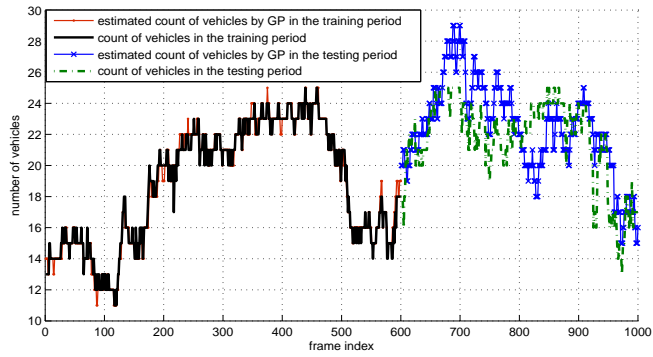}}
\end{tabular}
\caption{The first 600 frames are used for training the GP model and the last 400 frames is used for testing the accuracy of the traffic flow predicted by the GP.}\label{fig03}
\end{figure*}

We then compared the proposed method with a benchmark approach based on Gaussian mixture model (GMM) \cite{stauffer1999adaptive,tian2005robust}. The GMM approach uses the foreground detector and blob analysis to detect and count cars in the video sequence. The GMM is initialized by a certain number of video frames. After the training, the foreground detector begins to output more reliable segmentation results. The morphological opening is used to remove noise in the foreground and to fill gaps in the detected objects. Then blob analysis \cite{cucchiara2003detecting} is used to find bounding boxes of each connected component corresponding to
 a moving vehicle. Blobs which contain fewer than 150 pixels are rejected. Finally, the number of bounding boxes corresponds to the number of vehicles estimated by the GMM approach. The counting result with respect to one example frame under testing is plotted in Fig.\ref{GMM_res}. It is shown that the GMM approach failed to detect some vehicles and mistakenly recognized a single vehicle as several ones. Therefore its estimate on the number of vehicles is inaccurate. We selected four typical frames under testing to compare the GMM approach with the proposed method. The result is summarized in Table.\ref{Table:comparison}. It is shown that the proposed method provides much more accurate estimate than the GMM approach. In this comparative study, the  middle 400 video frames are reserved for training and the remaining frames are used for testing.
\begin{figure*}
\begin{tabular}{c}
\centerline{\includegraphics[width=2.5in,height=2in]{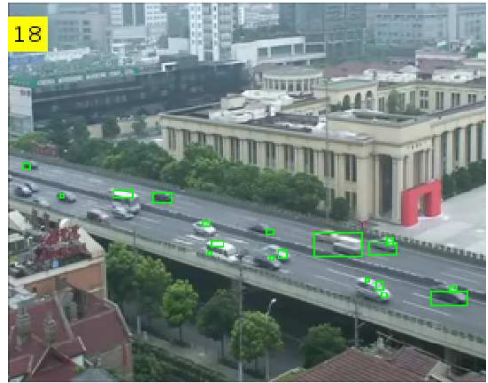}}
\end{tabular}
\caption{An example show of counting vehicles by the GMM approach. Each bounding box corresponds to a vehicle detected by the GMM approach. The number in the top left corner is the estimate given by the GMM approach.}\label{GMM_res}
\end{figure*}

\begin{table*}
\begin {center}
\begin{tabular}{c||c|c|c}
\hline Image index & true answer & estimate by GMM & estimate by the proposed method\\
\hline 262 & 21 & 12 & 20 \\
\hline 280 & 21 & 14 & 19 \\
\hline 752 & 23 & 14 & 20 \\
\hline 865 & 23 & 17 & 27 \\
\hline
\end{tabular}
\end {center}
\caption{Comparison of the GMM approach and our method in estimating the number of vehicles in four typical frames under testing.}\label{Table:comparison}
\end{table*}
\section{Conclusion}
In this paper, we present a vision based real-time TFP scheme using dynamic texture model for motion segmentation and GP for regression of counts of moving vehicles. We report work in progress in applying this scheme in real data analysis with low-resolution traffic visual images. Preliminary results show that this solution can give much more accurate estimate of the short-term TFP than a benchmark GMM approach. The current result is just preliminary, because the current version of the proposed scheme still has chance to be optimized. The GP parameters can be further optimized and many other features can be investigated so that we may discover more suitable features for use in TFP. The GP has been proven to be equivalent to a neural network with infinitely many hidden units \cite{neal2012bayesian}. In concept, the proposed TFP scheme can also be implemented by a deep GP, which has been recently proposed in \cite{damianou2013deep}. The relationship between deep GP and deep learning is worthy to be studied.
\section*{Acknowledgment}
This work was partly supported by the National Natural Science Foundation
(NSF) of China under grant Nos. 61571238, 61302158 and 61571434, the NSF of Jiangsu province under grant No. BK20130869 and the
China postdoctoral Science Foundation under grant Nos. 2015M580455 and 2016T90483.
\bibliographystyle{IEEEtran}
\bibliography{mybibfile}
\end{twocolumn}
\end{document}